\documentclass[runningheads]{llncs2}
\usepackage[T1]{fontenc}
\usepackage{graphicx}
\usepackage{url}
\usepackage{booktabs} 
\usepackage{multirow}
\usepackage{hyperref}
\usepackage[table,xcdraw]{xcolor}
\usepackage{amsmath}
\usepackage{amssymb}
\usepackage{mathtools}
\DeclareMathOperator*{\argmin}{arg\,min}
\usepackage{xcolor}
\usepackage{subcaption}

\newcommand{\AssumptionOneText}{For any $\psi(\cdot)$, $f(\cdot)$, there exists $f'(\cdot)$ such that its feature maps $\phi_{f'}(z)_i$ satisfy the constraints in \eqref{eq:scalar_product_constraint} for all $i,j\in\{1,\cdots, D\}$}

\newcommand{\LemmaOneText}{Under Assumption~\ref{eq:assumption}, the adversary $f'(\cdot)$ that satisfies \eqref{eq:scalar_product_constraint} is a global minimum of the minimization problem of \eqref{eq:f'_loss}.}

\newcommand{\TheoremOneText}{Minimizers $\psi(\cdot)$ of the minimization problem of \eqref{eq:task_source},~\ref{eq:f'_loss},~\ref{eq:psi_loss} align the marginal distributions of features in source and target domains \textit{i.e.}, $p(z)=q(z)$  at optimum.}

\begin{document}
\title{Unsupervised Adaptation of 3D CT Foundation Models for 3D CBCT Segmentation}
\author{
Gauthier Miralles\inst{1,2} \and
Loic Le Folgoc\inst{1} \and
Vincent Jugnon\inst{2} \and
Pietro Gori\inst{1}
}
\index{Miralles, Gauthier}
\index{Le Folgoc, Loic}
\index{Jugnon, Vincent}
\index{Gori, Pietro}

\authorrunning{G. Miralles et al.}

\institute{
LTCI, Télécom Paris, Institut Polytechnique de Paris, Palaiseau, France
\and
GE HealthCare, Buc, France
\\
\email{gauthier.miralles@ip-paris.fr}
}
\maketitle              
\begin{abstract}
Accurate 3D segmentation of cone-beam CT (CBCT) is critical for interventional and radiation therapy applications, yet it remains  limited by two compounding challenges: the scarcity of annotated CBCT data and the large domain shift from diagnostic CT. Interventional CBCT exhibits fundamental modality differences from conventional CT, driven by acquisition and physics effects as well as contrast-specific vascular content, thereby limiting effective cross-modality model transfer.
We propose a novel unsupervised domain adaptation (UDA) framework based on redundancy-reducing feature alignment, enabling 3D CBCT segmentation with no target-domain annotations or inference-time adaptation. Our framework is architecture-agnostic, seamlessly adapting both CNN-based and ViT-based foundation models (FMs). We evaluate our method on two challenging CT–CBCT liver segmentation benchmarks: one for interventional vascular procedures and one for radiation therapy, demonstrating that even large-scale pretrained segmentation networks require explicit feature-space bridging to generalize across acquisition modalities, and that our approach consistently outperforms existing pretrained FM and UDA strategies.
To support reproducibility and benchmarking, we release the liver segmentations for a public CBCT dataset, along with the code, trained models, and weights at \footnote{\url{https://github.com/mirabll/FARR3D/}}.

\end{abstract}
\section{Introduction}

Deep learning has become the standard for medical image segmentation, but it relies on large annotated datasets and assumes that training and deployment data follow the same distribution. When this assumption is violated, domain shift can severely degrade performance and limit clinical applicability. This issue is particularly pronounced for Cone-Beam CT (CBCT). Although CBCT resembles conventional CT, it differs substantially due to its limited reconstructed field of view, acquisition-geometry artifacts, physics-related degradations such as scatter and beam hardening, and locally injected iodine contrast producing high-intensity regions in interventional settings (Figure~\ref{fig:ct_cbct}). The domain gap between CT and CBCT restricts the transferability of pre-trained 3D CT foundation models.
Preprocessing pipelines designed for full-field CT often fail on CBCT, leading to unreliable zero-shot segmentation. Moreover, generative domain adaptation methods based on, for instance, CycleGAN~\cite{cyclegan,barbera2022,cycada} are not directly applicable, as they typically assume similar fields of view between source and target domains.\\
Existing 3D foundation models can be categorized by their level of automation. Fully automatic models include CNN-based approaches such as Merlin \cite{merlin} and ViT-based models such as MA-SAM \cite{masam}, whereas TotalSegmentator \cite{totalsegmentator}, despite being automatic, comprises multiple task-specific nnU-Net \cite{nnunet} models rather than a unified foundation model. Prompt-based models, inspired by Segment Anything \cite{sam} and predominantly ViT-based, include SAM-Med3D \cite{sammed3d} and MedSAM2 \cite{medsam2}. Hybrid models, such as Vista-3D, support both automatic inference and prompt-guided refinement \cite{vista3d}.\\
Given that CT is the closest available modality to CBCT, it serves as a natural source domain for adaptation. However, acquiring CBCT annotations is 
extremely time-consuming for segmentation, limiting the feasibility of supervised fine-tuning. We thus focus on unsupervised domain adaptation (UDA), where only labeled CT data are available, as a natural framework for CT-to-CBCT segmentation.
Restricting our comparison to methods operating natively in 3D, existing UDA approaches fall into three categories. Feature-alignment methods align source and target distributions in a shared latent space, either via adversarial discrimination, as in DA-nnUNet \cite{da_nnunet}, or through margin disparity discrepancy, as in MDD-UNet \cite{mdd_unet}, though the latter remains restricted to a specific segmentation task. Joint feature- and image-alignment methods, such as SIFA-3D \cite{sifa}, operate both in pixel and feature spaces, but require multiple GAN components and assume comparable fields of view between source and target, an unrealistic assumption for CT and CBCT. Self-training methods instead generate pseudo-labels in the target domain to supervise adaptation. As an example, MAPSeg \cite{mapseg} first leverages masked autoencoder pre-training to obtain a strong generalizable encoder, then applies pseudo-labelling to learn the segmentation task while preserving generalization ability.\\
In this work, we make minimal assumptions: no prior knowledge of field of view, no specialized initialization, and a lightweight design suitable for 3D volumes. To this end, we adopt a feature-level alignment strategy that works with any architecture, both CNN/ViT custom networks and foundation models.\\
\textbf{Contributions.} Our contributions are: (i) a 3D feature-level adaptation framework for adapting foundation model representations to CT-to-CBCT segmentation, applicable to CNN and ViT architectures; (ii) an evaluation on two complementary CT--CBCT datasets, covering both interventional and radiation therapy CBCT, demonstrating consistent gains over existing UDA methods; and (iii) the release of manually curated liver segmentations for the publicly available Pancreatic 3D CBCT dataset~\cite{hong2021pancreatic}, to facilitate future research and fair comparison.
\begin{figure}[htb]
\centering

\includegraphics[width=0.75\linewidth]{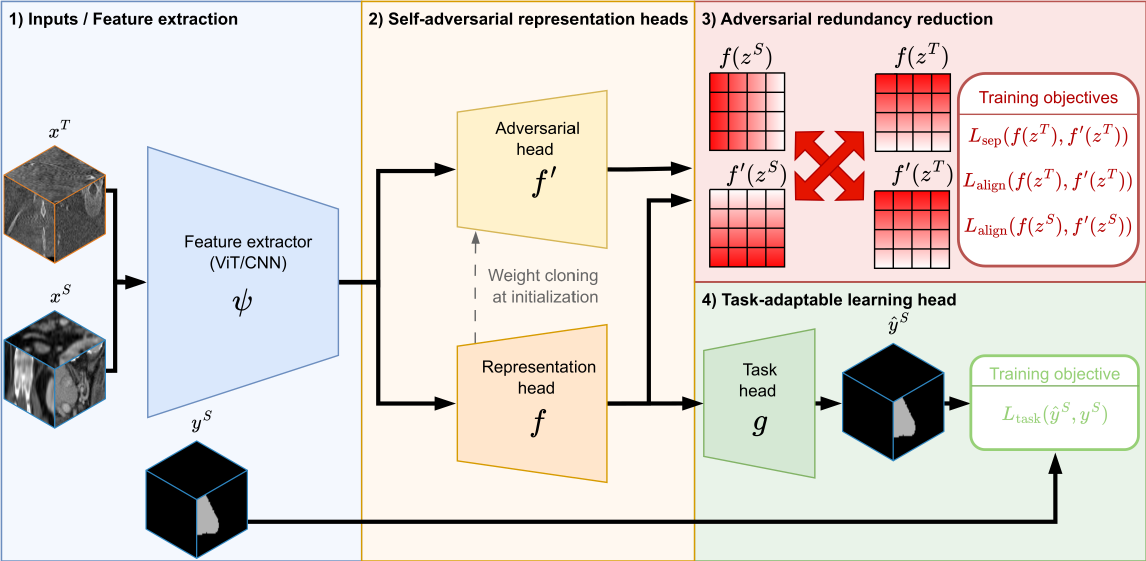}
\caption{Overview of the proposed UDA framework. Independently of the underlying backbone architecture, which can be either CNN- or ViT-based, the original network is decomposed into a shared feature extractor $\psi$, a representation head $f$, its adversarial counterpart $f'$ 
(i.e., a copy of $f$ at initialization), and a task-specific prediction head $g$. Given source-domain inputs $x^S$ (3D CT patches) and target-domain inputs $x^T$ (3D CBCT patches), the extracted features $z=\psi(x)$ are encouraged to become domain-invariant through a redundancy-reduction adversarial strategy acting between $f$ and $f'$. The prediction head $g$ is kept separate to allow adaptation to the target task, while $f'$ is discarded at inference time.}
\label{fig:framework}
\end{figure}

\section{Method}

Driven by the previous discussion, we aim to design a method that is generic, making no prior assumption on the type of domain shift, flexible, imposing no constraints on network architecture, and lightweight, allowing efficient training on 3D volumes with minimal computational resources. 

 We consider a source domain $\mathcal{D}_S = (\mathcal{X}^S, \mathcal{Y}, p^S(x, y))$ and a target domain $\mathcal{D}_T = (\mathcal{X}^T, \mathcal{Y}, p^T(x, y))$ with some domain shift captured by differing distributions $p^S$ and $p^T$. We are given labeled samples $(x_i^S, y_i^S)_{i=1}^{N_S}\sim p^S(x,y)$ in the source domain, and unlabeled samples $(x_i^T)_{i=1}^{N_T}\sim p^T(x)$ in the target domain. The goal is to learn a map $h:\mathcal{X}^S\cup \mathcal{X}^T\rightarrow \mathcal{Y}$ that produces a consistent labeling $h(x)$ for both source and target domain samples $x\in \mathcal{X}^S\cup \mathcal{X}^T$. We proceed via a feature-alignment strategy, whereby~(a) learned representations of source and target domain samples are driven to the same marginal distribution at convergence, 
and~(b) the map $h$ is made informative for the labeling task through explicit label supervision in the source domain. Next, we describe an adversarial strategy that leads to these two properties both theoretically and empirically. To derive the proposed approach, we split the predictor $h(x) \triangleq g(f(\psi(x)))$ into a feature encoder $\psi(\cdot)$, a representation head $f(\cdot)$ and a task-specific prediction head $g(\cdot)$. The resulting diagram in Figure~\ref{fig:framework} illustrates the workflow of the framework, showing how features are extracted, aligned, and made domain-invariant, while the task head $g$ remains separate for adaptation to the target task.\\
\textbf{Source domain label supervision.} The representation head $f(\cdot)$ and the prediction head $g(\cdot)$ are trained to optimally solve the labeling task in the source domain via a task-specific loss $\mathcal{L}_{\text{task}}$, as shown for one sample in Eq.\eqref{eq:task_source}:
\begin{equation}
    \argmin_{f,g} \, \mathcal{L}_{\text{task}}(h(x^S),y^S)
    \label{eq:task_source}
\end{equation}
\textbf{Adversarial feature alignment.} 
The features $z\triangleq \psi(x)$ are encouraged to become invariant to the 
domains, by introducing an adversarial strategy between $\psi(\cdot)$ and an adversary $f'(\cdot)$. The adversary $f'(\cdot)$ is trained to produce representations as dissimilar as possible from $f(\cdot)$ when using target samples $x^T$, 
and at the same time it should result in representations as similar as possible to $f(\cdot)$ when using source samples $x^S$. This is achieved in Eq.\eqref{eq:f'_loss} via a separation loss $\mathcal{L}_{\text{sep}}$ and an alignment loss $\mathcal{L}_{\text{align}}$, respectively, described in detail below. On the other hand, the feature extractor $\psi(\cdot)$ is trained with Eq.\eqref{eq:psi_loss} to output features $z\triangleq \psi(x)$ that~(1) are relevant for the labeling task; and~(2) that make the representations of the adversary $f'(z)$ and of the representation head $f(z)$ undistinguishable both in source and target domains:
\begin{equation}
    \argmin_{f'} \mathcal{L}_{\text{align}}\left(f(z^S),f'(z^S)\right) + \gamma \mathcal{L}_{\text{sep}}\left(f(z^T),f'(z^T)\right)
    \label{eq:f'_loss}
\end{equation}
\begin{equation}
    \argmin_{\psi}  \mathcal{L}_{\text{task}}(h(x^S),y^S) + \alpha \mathcal{L}_{\text{align}}\left(f(z^S),f'(z^S)\right) + \gamma \mathcal{L}_{\text{align}}\left(f(z^T),f'(z^T)\right)  
    \label{eq:psi_loss}
\end{equation}
\noindent where $(\alpha, \gamma) \in \mathbb{R}^{+} \times \mathbb{R}^{+}$ are user-defined hyperparameters.\\
\textbf{Algorithm.} In practice, we alternate between single optimization steps of \eqref{eq:task_source}, \eqref{eq:f'_loss} and \eqref{eq:psi_loss} using a stochastic gradient descent on source and target  batches. 

\noindent \textbf{Redundancy-reduction based representation alignment.}
The losses  $\mathcal{L}_{\text{align}}$ and $\mathcal{L}_{\text{sep}}$ should encourage alignment and separation of the latent representations $f(z),f'(z)$, respectively. To this end, we revisit Barlow Twins' redundancy reduction mechanism \cite{barlowtwins} and encourage the correlation matrix of representations $f(z),f'(z)$ to have specific structure conducive of these properties. Consider the representations $f(z)\in \mathbb{R}^{B\times D \times D_s \times H\times W}$ obtained from a batch. Let $\phi_f(z)\in \mathbb{R}^{D\times N}$ be the reshaped, flattened, centered and $L_2$-normalized (along $B,D_s,H,W$) version of $f(z)$, and similarly $\phi_{f'}(z)$ for $f'(z)$. Following~\cite{barlowtwins}, we note $\mathcal{C}[i,j]\triangleq \langle \phi_f(z)_i, \phi_{f'}(z)_j \rangle$ the cross-correlation of features $i,j\in \{1,\cdots, D\}$ for representations $f(z),f'(z)$. Let $p(z) \triangleq \psi_\sharp p^S(z)$ and $q(z) \triangleq \psi_\sharp p^T(z)$ denote the pushforwards of $p^S$ and $p^T$ through $\psi$, respectively.\\
\textbf{Alignment loss.} We define $\mathcal{L}_{\text{align}}\left(f(z),f'(z)\right)$ in Eq.\eqref{eq:L_align} as to encourage perfect correlation between the $f(\cdot)$ and the adversary $f'(\cdot)$ along homologous feature dimensions, and to discourage redundancy across different feature dimensions:
\begin{equation}
\mathcal{L}_{\text{align}}\left(f(z),f'(z)\right) \triangleq \sum_{i=1}^D (1 - \mathcal{C}[i,i])^2 + \frac{1}{D} \sum_{i\neq j} \mathcal{C}[i,j] ^2
    \label{eq:L_align}
\end{equation}
\textbf{Separation loss.} Differently, we define $\mathcal{L}_{\text{sep}}\left(f(z),f'(z)\right)$ in Eq.\eqref{eq:L_sep} to encourage perfect decorrelation between $f(\cdot)$ and $f'(\cdot)$ across all feature dimensions:
\begin{equation}
    \mathcal{L}_{\text{sep}}\left(f(z),f'(z)\right) \triangleq \sum_{i=1}^D \mathcal{C}[i,i]^2 + \frac{1}{D} \sum_{i\neq j} \mathcal{C}[i,j] ^2
    \label{eq:L_sep}
\end{equation}
 $\mathcal{L}_{\text{align}}$ and $\mathcal{L}_{\text{sep}}$ are task-agnostic: they encourage representations to match or differ without being specific to a task ( \textit{i.e.}, $g(\cdot)$). This contrasts with previous works, such as MDD \cite{mdd_unet}, where specific losses, such as cross-entropy for classification, were used. Furthermore, by using the proposed redundancy-reduction losses, we do not need an adversary task-specific prediction head $g'(\cdot)$, as in MDD, since we move the feature alignment from the space of $g$ to the one of $f$.

\noindent \textbf{Theoretical guarantee.} We rewrite the adversarial loss in terms of the correlation matrix 
$w(z)_{i,j} = \langle \phi_f(z)_i , \phi_{f'}(z)_j \rangle$.
Since the resulting relaxed objective is separable and quadratic in each 
$w(z)_{i,j}$, its minimizer is:

\begin{equation}\label{eq:scalar_product_constraint}
w(z)_{i,j} =
\begin{cases}
\dfrac{p(z)}{p(z) + \gamma q(z)} & \text{if } i=j,\\[4pt]
0 & \text{if } i\neq j.
\end{cases}
\end{equation}

\noindent Since this solution may not be realizable by any $f'$, we introduce a
capacity assumption, similarly to 
MDD,
which relies on sufficiently expressive classifiers to estimate divergence and
ensure theoretical feature alignment.
\newtheorem{assumption}{Assumption}
\begin{assumption}
\AssumptionOneText
\label{eq:assumption}
\end{assumption}

\begin{lemma}
\LemmaOneText
\label{le:lemma}
\end{lemma}

\begin{theorem}\label{th:theorem}
\TheoremOneText
\end{theorem}

\paragraph{Proof sketch.}
Using Lemma~\ref{le:lemma}, we substitute the optimal $f'$ into the $\psi$-objective and minimize the resulting Lagrangian with respect to $\psi$ (through the induced density $q=\psi_\sharp p^T$), under the normalization constraint $\int q(z)\, dz = 1$. First-order optimality conditions then imply $p(z)=q(z)$ at optimum.

\section{Experiments and Results}
We evaluated our method on 3D liver segmentation using two CT--CBCT datasets. 
The first is an unpaired CT--CBCT dataset constructed from public sources: 
Pancreatic-CT-CBCT-SEG~\cite{hong2021pancreatic} (40 paired scans) and LiTS~\cite{lits} (130 CT scans with liver annotations). Since Pancreatic-CT-CBCT-SEG 
lacks liver masks, we generated and manually curated annotations on the CT volumes 
and transferred them to CBCT via the registered scans. The final dataset contains 130 CT (LiTS) and 39 CBCT (DR) volumes and will be publicly released for reproducibility. 
The second dataset is a private clinical collection of 678 CT and 573 interventional CBCT (DI) volumes with manually curated liver segmentations.
For both datasets, splits were performed at the patient level, with 2/3 of the patients used for training and validation and the remaining 1/3 reserved for testing.

\noindent \textbf{Experimental details.}
All CT and CBCT volumes are resampled to isotropic 1.8mm voxel spacing. All UDA methods share the same 3D U-Net backbone trained with standard augmentations. CNN- and ViT-based foundation models are evaluated with their official implementations. Prompted variants use 1 or 10 randomly sampled liver points.
For nnUNet and VISTA-3D, the representation head $f$ corresponds to the penultimate stage, and for SAM-Med3D to the final 3D attention block. In both cases, $f'$ is initialized as a copy of $f$, $\psi$ comprises all preceding stages, and $g$ all subsequent ones (Figure~\ref{fig:framework}). Full implementation details are available in our \href{https://github.com/mirabll/FARR3D/}{code repository}.

\noindent \textbf{Baselines.} We compare our method against state-of-the-art UDA approaches for 3D CBCT segmentation without target-domain annotations.
These include zero-shot inference using pretrained foundation models on CBCT data, both in automatic and prompt-based mode \cite{merlin,totalsegmentator,masam,vista3d,sammed3d,medsam2}.
We also consider 3D unsupervised domain adaptation strategies that exploit  CT data to bridge the CT–CBCT domain gap through feature alignment \cite{da_nnunet,mdd_unet}, image translation \cite{sifa}, or self-training \cite{mapseg}.
All methods are evaluated under identical training and evaluation protocols to ensure a fair comparison.

\noindent
\begin{minipage}[t]{0.49\linewidth}
\centering
\includegraphics[width=\linewidth]{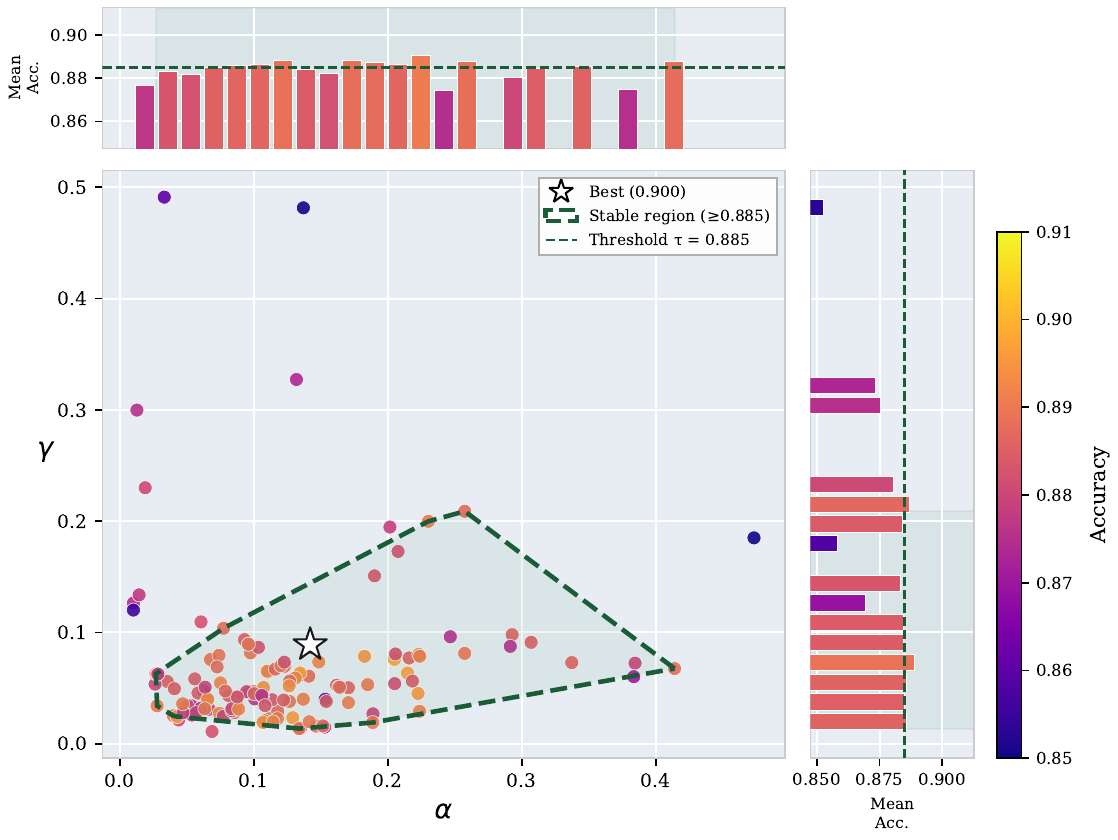}
\captionof{figure}{Sensitivity to hyperparameters $\alpha$ and $\gamma$ on dataset DR. Each point corresponds to an independent run (colored by F1, \%). The polygon shows points with F1 $\geq$ 88.5\%. The star ($\bigstar$) marks the best result (F1 $=$ 90.0\%).}
\label{fig:stability}
\end{minipage}
\hfill
\begin{minipage}[t]{0.49\linewidth}
\centering
\includegraphics[width=\linewidth]{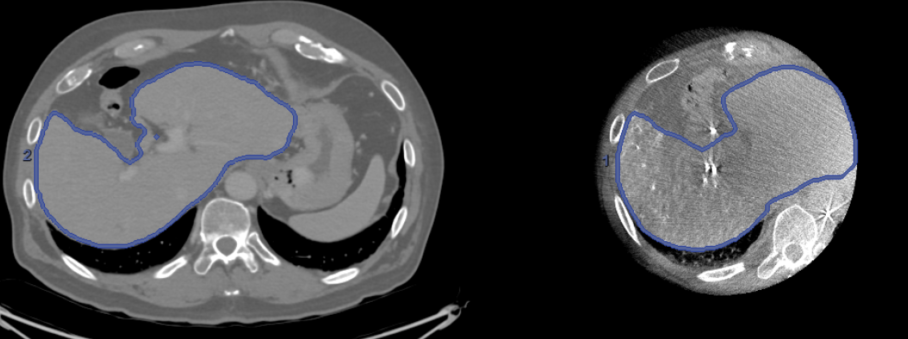}
\captionof{figure}{Qualitative comparison of CT and CBCT. Axial slices of the same patient displayed with identical visualization windows. The liver contour (blue) is overlaid on CT (left) and CBCT (right).}
\label{fig:ct_cbct}
\end{minipage}

\noindent \textbf{Unsupervised Domain Adaptation.} Table~\ref{tab:results} benchmarks recent 3D UDA methods using F1 score across two CT–CBCT datasets. Image-alignment approaches such as SIFA-3D underperform due to the large field-of-view discrepancy between domains, while MAPSeg fails to surpass the source-only baseline, likely because its masked autoencoder pre-training combined with pseudo-labeling is sensitive to poor initialization, causing error propagation during self-training. Feature-alignment strategies prove more effective and lightweight overall, with DA-nnUNet consistently outperforming MDD-UNet on both datasets. Our method achieves the best results across all benchmarks while remaining lightweight, making it well-suited for integration with 3D foundation models. Furthermore,
an ablation study on the sensitivity to hyperparameters $\alpha$ and $\gamma$ (Figure~\ref{fig:stability})
demonstrates that performance remains stable over a wide range of values, with
a large contiguous region achieving near-optimal F1 scores. These results indicate that our approach is robust to hyperparameter selection.

\begin{table}[t]
\caption{F1 performance evaluation of UDA strategies and zero-shot foundation models for CBCT liver segmentation on two datasets. Source: CT, Target: radiation therapy CBCT (DR) or interventional CBCT (DI). Best results (per category) in \textbf{bold}. $\Sigma_p$ (M) denotes the total number of additional trainable parameters introduced by the UDA method (in millions). }
\label{tab:results}
\centering
\begin{tabular}{llccc|cc}
\toprule
{\textbf{Type}} & \textbf{Method} & \textbf{Year} & \textbf{Mode} & \multicolumn{2}{c}{\textbf{F1} \( \uparrow \)} & \textbf{$\Sigma_p$}(M) \( \downarrow \) \\
\cmidrule(lr){5-6}
& & & & \textbf{DR} & \textbf{DI} & \\
\midrule
{\centering \textbf{Source Only}} & nnUNet \cite{nnunet} & & & 66.5 & 80.1 & \\
\midrule
\multirow{5}{*}{\parbox[t]{2.5cm}{\centering \textbf{Unsupervised}\\\textbf{Domain Adaptation}}} & DA-nnUNet \cite{da_nnunet} & 2024 & Auto & 73.3 & 84.6 & 6.62 \\
& MDD-UNet \cite{mdd_unet} & 2023 & Auto & 66.7 & 80.6 & 0.09 \\
& SIFA-3D \cite{sifa} & 2023 & Auto & 55.2 & 64.7 & 37.69 \\
& MAPSeg \cite{mapseg} & 2024 & Auto & 59.9 & 70.2 & 210.96 \\
& \cellcolor{gray!12}nnUNet + Ours & \cellcolor{gray!12}2026 & \cellcolor{gray!12}Auto & \cellcolor{gray!12}\textbf{78.0} & \cellcolor{gray!12}\textbf{86.0} & \cellcolor{gray!12}\textbf{0.08} \\
\midrule
\multirow{12}{*}{\parbox[t]{2.5cm}{\centering \textbf{Foundation}\\\textbf{Models}}} & MA-SAM \cite{masam} & 2024 & Auto & 15.4 & 61.8 & \\
& Merlin \cite{merlin} & 2025 & Auto & 18.8 & 33.7 & \\
& TotalSegmentator \cite{totalsegmentator} & 2023 & Auto & 64.1 & 73.8 & \\
\cmidrule(r){2-7}
& \multirow{2}{*}{MedSAM2 \cite{medsam2}} & \multirow{2}{*}{2025} & 1 pt & 40.4 & 30.7 & \\
& & & 10 pts & 69.7 & 64.6 & \\
\cmidrule(r){2-7}
& \multirow{2}{*}{SAM-Med3D \cite{sammed3d}} & \multirow{2}{*}{2023} & 1 pt & 44.2 & 53.6 & \\
& & & 10 pts & 73.8 & 65.3 & \\
& \cellcolor{gray!12}SAM-Med3D + Ours & \cellcolor{gray!12} & \cellcolor{gray!12}10 pts & \cellcolor{gray!12}77.6 & 74.0\cellcolor{gray!12} & \cellcolor{gray!12}11.37 \\
\cmidrule(r){2-7}
& \multirow{3}{*}{VISTA-3D \cite{vista3d}} & \multirow{3}{*}{2025} & Auto & 42.7 & 48.3 & \\
& & & 1 pt & 58.4 & 48.9 & \\
& & & 10 pts & 80.0 & 74.9 & \\
& \cellcolor{gray!12}VISTA-3D + Ours & \cellcolor{gray!12} & \cellcolor{gray!12}10 pts & \cellcolor{gray!12}\textbf{90.0} & \cellcolor{gray!12}\textbf{81.5} & \cellcolor{gray!12}0.99 \\
\midrule
{\centering \textbf{Target Only}} & nnUNet \cite{nnunet} & & & 91.4 & 93.7 & \\
\bottomrule
\end{tabular}
\end{table}

\begin{figure*}[htb]
    \centering
    \includegraphics[width=0.68\textwidth]{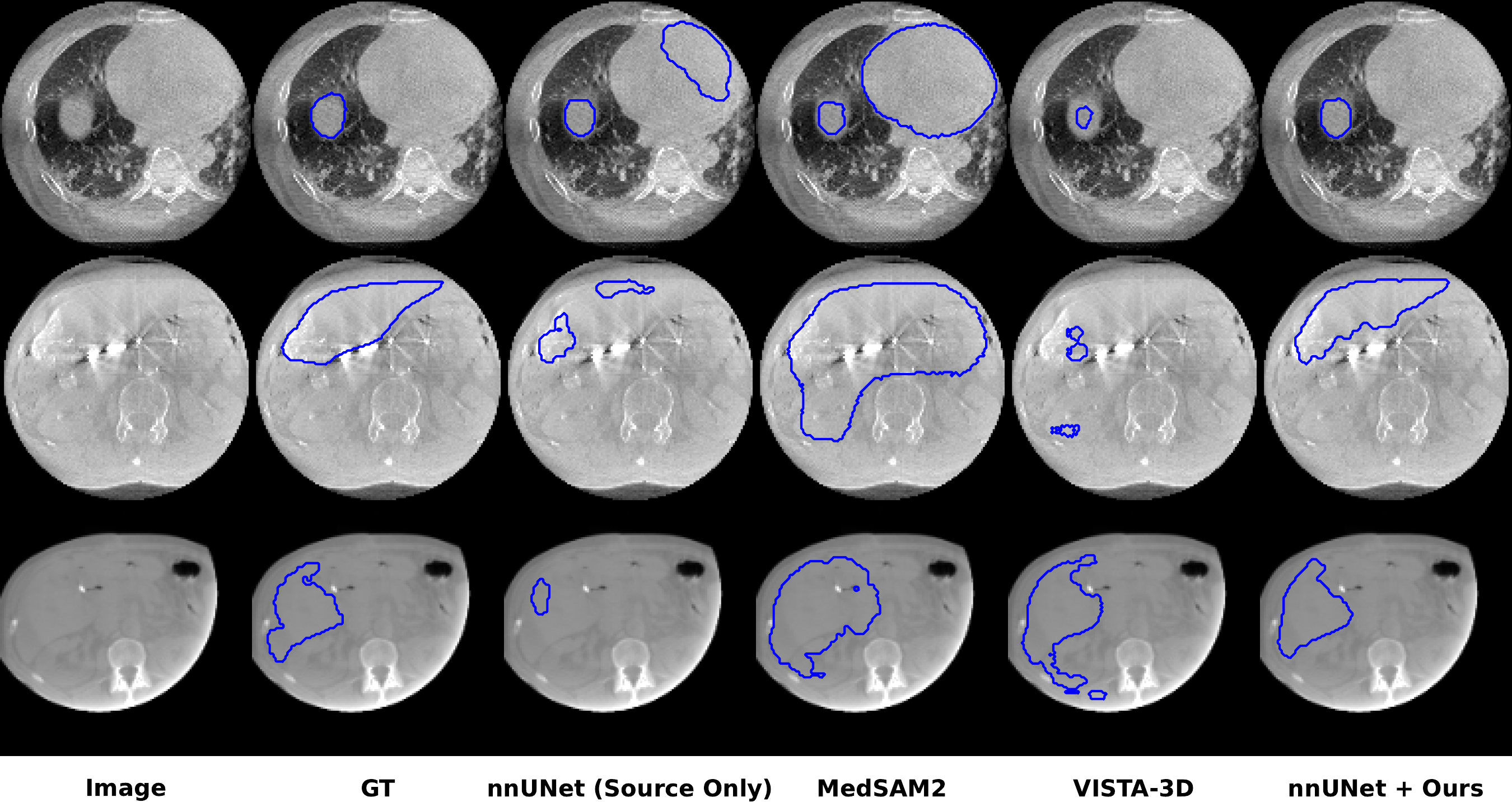}
    \caption{
    Qualitative comparison of liver segmentation on the DI (top two rows) and DR (bottom row) test sets.
    Left to right: input image, ground truth (GT), source-only model,
    MedSAM2 (zero-shot, 10 prompts), VISTA-3D (zero-shot, 10 prompts), and
    our UDA-adapted model. Predicted masks are shown in blue.
    }
    \label{fig:liver_comparison}
\end{figure*}

\noindent \textbf{Foundation Models.} In contrast to UDA methods, several foundation models, including VISTA-3D (interactive branch) and MedSAM2, perform worse on the DI dataset than on DR. This likely stems from iodine-induced high intensities, which fall outside the range expected by the models' default preprocessing, such as min–max normalization or CT-specific intensity scaling. As a result, these models operate on degraded inputs, whereas nnUNet benefits from percentile normalization (0.5th–99.5th), which clips such intensity outliers.
Overall, most 3D foundation models fail to achieve satisfactory zero-shot liver segmentation on CBCT data in automatic mode. When using point prompts (10 points), SAM-Med3D and VISTA-3D provide the strongest initializations. Our method enables the adaptation of both ViT-based (SAM-Med3D) and CNN-based (VISTA-3D) architectures, leading to consistent and substantial performance gains without requiring any labeled target-domain data. Qualitative results on both datasets (Figure~\ref{fig:liver_comparison}) indicate that MedSAM2 tends to over-segment and exhibits reduced sensitivity to high-intensity structures compared to VISTA-3D. After adaptation with our UDA framework, the source-only model becomes more robust to high-intensity artifacts (Figure~\ref{fig:liver_comparison}, row 2) and field-of-view variations (Figure~\ref{fig:liver_comparison}, row 3), leading to more anatomically consistent liver segmentations.

\section{Conclusion}
We investigate unsupervised adaptation of 3D CT foundation models for 3D CBCT segmentation under severe cross-modality domain shift. While 3D foundation models can provide strong structural priors, direct transfer to CBCT is insufficient without explicit feature-space adaptation. By introducing a redundancy-reducing alignment strategy, pretrained models can be effectively bridged across modalities, improving performance on both interventional vascular and radiation therapy CT–CBCT benchmarks. Our framework is architecture-agnostic, supports both CNN- and ViT-based models, and requires no target-domain annotations or inference-time optimization, making it clinically practical.

Although demonstrated on liver segmentation, the approach generalizes to multi-organ tasks and may also benefit other label-scarce interventional imaging applications, such as artifact correction, denoising, or pose estimation, where domain shift limits performance. We believe this work highlights the importance of explicit representation alignment for unlocking the full potential of foundation models in CBCT.

\subsubsection*{Acknowledgements.}
This work was partially funded by the French Ministry for Higher Education and Research as part of CIFRE grant No.~2024/0313 and supported by GE~HealthCare, where VJ and GM are employed. This work was performed using HPC resources from GENCI-IDRIS (Grant 200319327Z and 2024-A0160615058).

\subsubsection*{Disclosure of Interests.}
G.\,Miralles and V.\,Jugnon are employed by GE~HealthCare. The remaining authors have no competing interests to declare that are relevant to the content of this article.

\clearpage \appendix \begin{center} {\Large\bfseries Supplementary Material} \end{center} \vspace{1em} \section{Reproducibility and Model Integration} \label{app:reproducibility} \noindent\textbf{Public dataset release.} We release liver segmentation masks for the 39 paired CT--CBCT cases of Pancreatic-CT-CBCT-SEG. The original collection provides expert annotations for several organs at risk but does not contain liver masks. We contribute paired CT and CBCT liver masks in NIfTI format, using the naming convention \texttt{LiverCT\_XXXX.nii.gz} and \texttt{LiverCBCT\_XXXX.nii.gz}, respectively. The complete paired dataset, including the images and released masks, is publicly available through the repository referenced in the main paper. 

\medskip \noindent\textbf{Dataset organization and preprocessing.} Source CT and target CBCT volumes are organized using separate \texttt{imagesTr}, \texttt{imagesTs}, \texttt{labelsTr}, and \texttt{labelsTs} directories. All images and label maps are resampled to isotropic $1.8$\,mm voxel spacing. Image volumes are resampled using continuous interpolation, whereas nearest-neighbor interpolation is used for segmentation masks to preserve their discrete labels. Training and test partitions are defined at the patient level. 

\medskip \noindent\textbf{Common backbone for UDA comparisons.} For comparisons between feature-alignment methods, we use the same 3D U-Net backbone comprising five resolution stages, 64 base channels, two $3\times3\times3$ convolutions per stage, LeakyReLU activations, and skip connections. Standard spatial and intensity augmentations include rotation, scaling, flipping, contrast modification, and additive noise. This common configuration isolates the effect of the adaptation objective from differences in backbone architecture or preprocessing. \medskip 

\noindent\textbf{Integration into VISTA-3D.} To integrate our method into VISTA-3D, the network is decomposed into a feature extractor $\psi$, a representation head $f$, and the subsequent task head $g$. An adversarial head $f'$ is initialized as a copy of $f$ and is optimized only during adaptation. Our implementation introduces an adapted VISTA-3D class containing these components and the redundancy-reduction objectives. After training, $f'$ is removed, and inference uses the original feature and task paths. Consequently, adaptation introduces no additional inference-time module or optimization.

\medskip \noindent\textbf{Training and evaluation workflow.} All paths, adaptation coefficients, and training options are specified in a single configuration file. The released implementation provides separate scripts for adaptation and target-domain evaluation, together with the source-only and adapted checkpoints. The repository also provides the data loading, preprocessing, validation, learning-rate scheduling, and metric computation procedures used for the reported experiments.

\end{document}